\pdfoutput=1
\documentclass{article}
\usepackage[preprint]{neurips_2023}
\usepackage{graphicx} % Required for inserting images
\usepackage{parskip}
\usepackage{amsmath}
\usepackage{amsfonts}
\usepackage{hyperref}
\usepackage{cleveref}
\usepackage{appendix}
\usepackage{float}
\usepackage{subcaption}
\usepackage{blindtext}
\usepackage{booktabs}
\usepackage{caption} 
\title{PairVDN - Pair-wise Decomposed Value Functions}
\author{%
  Zak Buzzard \\
  Department of Computer Science\\
University of Cambridge\\
  \texttt{zzb20@cam.ac.uk} \\
}

\DeclareMathOperator*{\argmax}{argmax}

\begin{document}

\maketitle

\newcommand{\bracket}[1]{\left(#1\right)}

\newcommand{\St}{\mathcal{S}}
\newcommand{\A}{\mathcal{A}}
\newcommand{\T}{\mathcal{T}}
\newcommand{\R}{\mathcal{R}}
\newcommand{\state}{s}
\newcommand{\ob}{o}
\newcommand{\act}{a}

\newcommand{\obs}{\ob_1,\dots,\ob_n}
\newcommand{\acts}{\act_1,\dots,\act_n}

\newcommand{\Q}[1]{\tilde{Q}_{#1}}
\newcommand{\Qq}[2]{\tilde{Q}_{#1,#2}}

\begin{abstract}
Extending deep Q-learning to cooperative multi-agent settings is challenging due to the exponential growth of the joint action space, the non-stationary environment, and the credit assignment problem. Value decomposition allows deep Q-learning to be applied at the joint agent level, at the cost of reduced expressivity. Building on past work in this direction, our paper proposes PairVDN, a novel method for decomposing the value function into a collection of pair-wise, rather than per-agent, functions, improving expressivity at the cost of requiring a more complex (but still efficient) dynamic programming maximisation algorithm. Our method enables the representation of value functions which cannot be expressed as a monotonic combination of per-agent functions, unlike past approaches such as VDN and QMIX. We implement a novel many-agent cooperative environment, Box Jump, and demonstrate improved performance over these baselines in this setting. We open-source our code and environment at \url{https://github.com/zzbuzzard/PairVDN}.
\end{abstract}

\section{Introduction}
Multi-agent reinforcement learning (MARL) deals with the application of reinforcement learning to settings with more than one agent, for instance team-based videogames, the coordination of multiple self-driving vehicles, or competitive games such as chess or go. However, the complexity of multiple agents brings various challenges for reinforcement learning: the size of the joint action space grows exponentially with the number of agents, quickly becoming intractable to enumerate; credit assignment may become difficult, particularly in environments with joint reward systems; the presence of other agents creates a non-stationary environment from the perspective of individual agents, who must now account for the behaviour of the other agents (which changes during optimisation). This paper examines the fully cooperative multi-agent setting, in which several agents jointly work for the same goal, with a single reward signal shared by all agents.

Deep Q-networks (DQN) \cite{DQN} are a method for learning policies via deep learning, and has seen success across a range of tasks. However, as DQN requires maximisation over the entire joint action space, this is not feasible in a multi-agent setting with more than a few agents. While it is possible to apply DQN individually to each agent - a naive approach called independent Q-learning (IQL) - the non-stationarity of the environment in this setting causes low performance \cite{QMIX}. To overcome these issues, Value Decomposition Networks (VDN) \cite{VDN} were proposed, which avoid the high cost of iterating over the joint action space by decomposing the joint value function $Q$ into a sum of per-agent value functions:
\begin{align*}
Q((o_1, \dots, o_n), (a_1, \dots, a_n)) \approx \sum_{i=1}^n \tilde{Q}_i(o_i, a_i)
\end{align*}
for $n$ agents with observations $o_1, \dots, o_n$ and actions $a_1, \dots, a_n$.
Due to the monotonic relationship between $\tilde{Q}_i$ and $Q$, maximising $Q$ is now tractable by the independent maximisation of each per-agent value function $\tilde{Q}_i$. QMIX extends this idea further with a more expressive decomposition function: rather than combine value functions via simple summation, they define a monotonic MLP (achieved through non-negative weights and an absolute value maximisation function) which processes the per-agent Q-values to produce the joint value. The weights for the MLP depend on the environment's state via a HyperNetwork \cite{HyperNetworks}. This is a more expressive parameterisation, allowing the representation of a larger set of joint Q functions. However, the constraint of monotonicity is limiting; each agent acts based only on its local Q-function $\tilde{Q}_i$, and there is no way to represent the strength of joint actions. For example, imagine several agents controlling cars at a crossroads, each with actions stop or go. In this scenario, the joint value of any two agents simultaneously going (and crashing into one another) is very low - this cannot be represented by either VDN or QMIX.

% Todo continue
This paper proposes an alternative \textit{non-monotonic} decomposition of the joint value function, and the accompanying non-trivial maximisation algorithm.

% IQL, VDN, QMIX?

\section{Paired VDN} \label{sec:paired_vdn}

Both VDN and QMIX are restricted to representing joint $Q$-functions which decompose into a monotonic combination of per-agent $Q$-functions $\Q{1},\dots,\Q{n}$, allowing the easy maximisation of $Q$ as
\begin{equation} \label{eq:cond}
\argmax_{\acts} Q((\obs), (\acts)) = \begin{pmatrix}
\argmax_{\act_1} \Q{1}(\ob_1, \act_1)\\
\vdots\\
\argmax_{\act_n} \Q{n}(\ob_n, \act_n)\\
\end{pmatrix},
\end{equation}
which also has the benefit of enabling decentralized execution: agent $i$ just greedily maximises $\Q{i}$. Decentralization aside, $\cref{eq:cond}$ enables the practical application of DQN to multi-agent settings with a large number of agents $n$, as it avoids the need to iterate directly over $\A^n$. In this project, we explore an alternative parameterisation of $Q$ which does \textit{not} satisfy $\cref{eq:cond}$, allowing greater expressivity, but nevertheless allows \textit{tractable} maximisation of $Q$ in sub-exponential time in the number of agents. In particular, consider a decomposition of $Q$ into $n$ pair-wise functions $\Qq{i}{j}$, a method we name PairVDN, such that
\begin{equation} \label{eq:pairvdn}
Q((\obs), (\acts)) \approx \sum_{i=1}^n \Qq{i}{j}((\ob_i, \ob_j), (\act_i, \act_j))
\end{equation}
where $j=i+1$ for $i<n$, wrapping around to $j=1$ at $i=n$ (technically $j$ is a function of $i$, but we leave this notation out for simplicity), and the indices $[1,2,\dots,n]$ denote some arbitrary ordering of the agents. Intuitively, $\Qq{i}{j}$ gives some indication of the effect on future rewards of agents $i$ and $j$ jointly taking actions $(\act_i, \act_j)$. The terms form a cycle over the agents, with each action accounted for by two terms, rather than one in VDN. PairVDN is strictly more expressive than VDN, as the network may learn $\Qq{i}{j}((\ob_i,\ob_j),(\act_i,\act_j))=\Q{i}(\ob_i,\act_i)$, but is neither more nor less expressive than QMIX.

Intuitively, our formulation allows agent actions to take into account those of other agents at the same timestep. For example, imagine an environment in which all agents must work together to break a wall, and all must simultaneously hit the wall for the best result. PairVDN is able to effectively represent this value function by assigning a positive weight to $\Qq{i}{j}((\ob_i,\ob_j),(\act_i,\act_j))$ only when $\act_i=\act_j$. This function is clearly maximised for $\act_1=\act_2=\dots=\act_n$, causing the agents to work together and act in unison as is required. This example value function does not satisfy \eqref{eq:pairvdn} for any set of per-agent Q-functions.

Practically, this may be implemented in a very similar manner to VDN, but with each neural network $\Qq{i}{j}$ receiving two observations as input and outputting a vector of length $|\A|^2$, which is then reshaped into an $|\A|\times|\A|$ grid. The $n$ networks $\Qq{1}{2},\dots\Qq{n}{1}$ may share parameters as with VDN and QMIX.

As each action $\act_i$ occurs in exactly two terms in \eqref{eq:pairvdn}, maximising $Q$ is no longer trivial, and may not be achieved in a decentralized manner. We achieve efficient maximisation through a dynamic programming algorithm, with time complexity $\mathcal{O}\left(n |\A|^3\right)$, scaling \textit{linearly} with the number of agents, although poorly with $|\A|$. This algorithm is asymptotically superior to naive iteration over the joint action space, which requires $\mathcal{O}\left(|\A|^n\right)$ time, and quickly becomes infeasible as $n$ grows. Full algorithm details may be found in \Cref{app:dynamic_programming}.

% However, it remains achievable: using dynamic programming, \eqref{eq:pairvdn} may be maximised in $\mathcal{O}\left(n |\A|^3\right)$ time - details in \Cref{app:dynamic_programming}. This method scales linearly with the number of agents, but poorly with $|\A|$.

\section{Experiments}

\subsection{Environments}

% Environments figure - probably has to go, or appendix it
\newcommand{\hh}{2.5cm}
\begin{figure}
  \centering
  \begin{subfigure}{0.25\linewidth}
    \centering
    \centerline{\includegraphics[height=\hh]{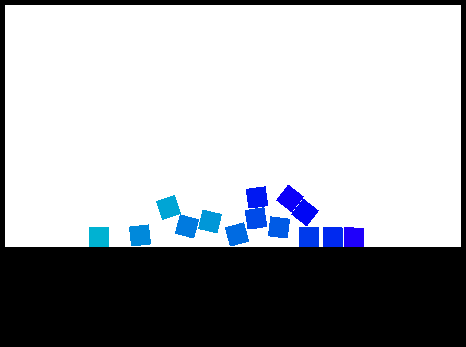}}
    \caption{Box Jump}
  \end{subfigure}
  \hfill
  \begin{subfigure}{0.25\linewidth}
    \centering
    \centerline{\includegraphics[height=\hh]{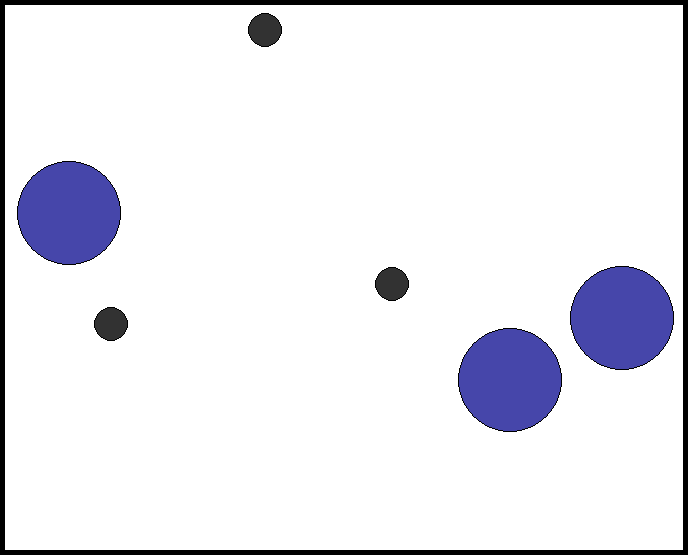}}
    \caption{Simple Spread}
  \end{subfigure}
  \hfill
  \begin{subfigure}{0.25\linewidth}
    \centering
    \centerline{\includegraphics[height=\hh]{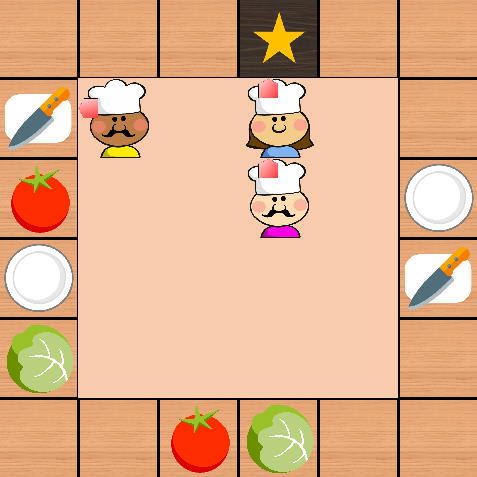}}
    \caption{CookingZoo}
  \end{subfigure}
  \caption{The three environments we evaluate on. Box Jump is a custom environment we design for this project.}
  \label{fig:envs}
\end{figure}

Limited to fully co-operative environments with discrete action spaces, with a preference for simple environments with support for many agents (and global state - required by QMIX as input to the hypernetworks), we select the following environments:
\begin{itemize}
\item Simple Spread: a simple inbuilt environment in which agents must get as close as possible to target locations without touching one another.
\item Cooking Zoo \cite{cooking_zoo}: a third-party environment inspired by the videogame Overcooked and adapted to PettingZoo. We modify this slightly to add support for global state and an easier level.
\item Box Jump: a custom environment we designed during the project. It is a relatively simple 2D physics-based environment, in which each agent is a box, and agents are rewarded based on the maximum height of any agent, incentivising them to work collaboratively to build a tower. It is designed with a large number of agents in mind, and runs well with 16 or more. We emphasise that PairVDN has no inherent advantage on Box Jump; it was not designed to favour our approach, but to allow scaling to many agents. Full details in \Cref{app:box_jump}.
\end{itemize}
The environments are shown in \Cref{fig:envs}. In each case, agents receive only a shared global reward signal. We stress that this makes each environment highly challenging, especially with a large number of agents, due to the credit assignment problem.

\subsection{Setup}
Each joint value function $\tilde{Q}_{i,j}((o_i, o_j), (\act_i, \act_j))$ is parameterised as a multi-layer perceptron with ReLU activation, input dimension $2|\St|$, and output dimension $|\A|^2$ giving the estimated Q-values for each pair of actions $(a_1,a_2)\in\A^2$. The parameters of the MLP are shared across all agents to accelerate training, with one-hot agent identifiers appended to the observations. Our codebase is implemented from scratch using the \verb|jax|, \verb|equinox| and \verb|optax| libraries.

As suggested in DQN \cite{DQN} we employ a modified objective based on a target network with parameters $\theta^-$. The target network is updated via the exponential moving average,
\[
    \theta^-_{t+1} = c \theta^-_t + (1-c) \theta
\]
following the $t_\text{th}$ update of the parameters $\theta$. We employ an $\epsilon$-greedy policy to encourage exploration, with the value of $\epsilon$ interpolated during training from $\epsilon=1$ to $\epsilon=0.05$. Full hyperparameter details may be found in \Cref{app:hyperparams}.

We compare PairVDN to VDN, QMIX, and independent Q-learning (IQL). We find it beneficial to initialise QMIX in a HyperNetwork-specific manner \cite{hypernetwork_init}.

\subsection{Results}

\Cref{fig:marl_reward_graphs} shows rewards during training for three Box Jump tasks, and the cooking and simple spread tasks. All models perform similarly on the 8 agent Box Jump task, but when the number increases to 16, QMIX becomes unstable, and IQL becomes ineffective, with the difference further exaggerated in the variant without box rotation.

\begin{figure}
  \centering
  \begin{subfigure}{0.22\linewidth}
    \centering
    \includegraphics[width=\linewidth, trim=0cm 0cm 0cm 6cm, clip]{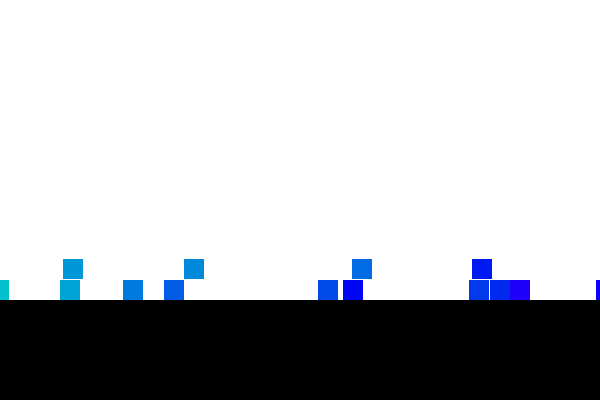}
    \caption{VDN, $t=400$}
  \end{subfigure}
  % \hfill
  \hspace{1pt}
  \begin{subfigure}{0.22\linewidth}
    \centering
    \includegraphics[width=\linewidth, trim=0cm 0cm 0cm 6cm, clip]{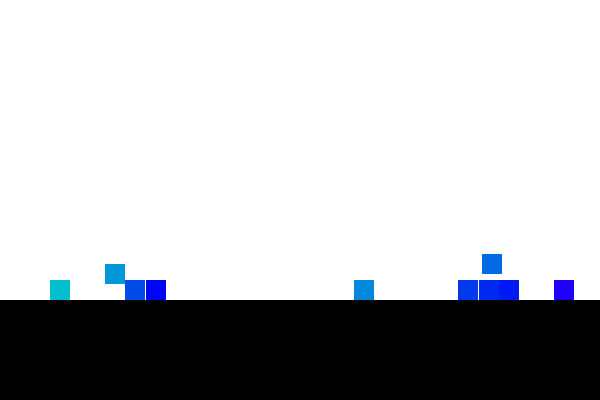}
    \caption{VDN, $t=1000$}
  \end{subfigure}
  \hfill
  \begin{subfigure}{0.22\linewidth}
    \centering
    \includegraphics[width=\linewidth, trim=0cm 0cm 0cm 6cm, clip]{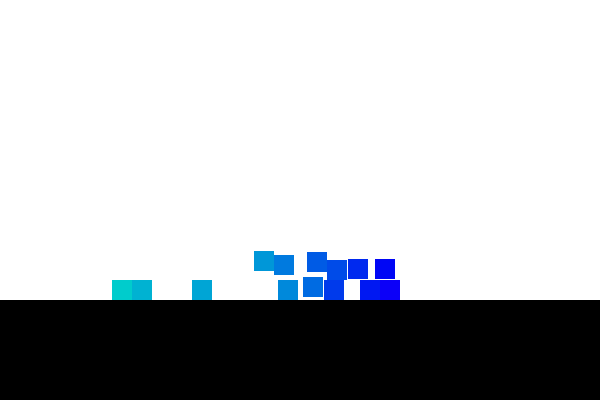}
    \caption{PairVDN, $t=400$}
  \end{subfigure}
  \hspace{1pt}
  \begin{subfigure}{0.22\linewidth}
    \centering
    \includegraphics[width=\linewidth, trim=0cm 0cm 0cm 6cm, clip]{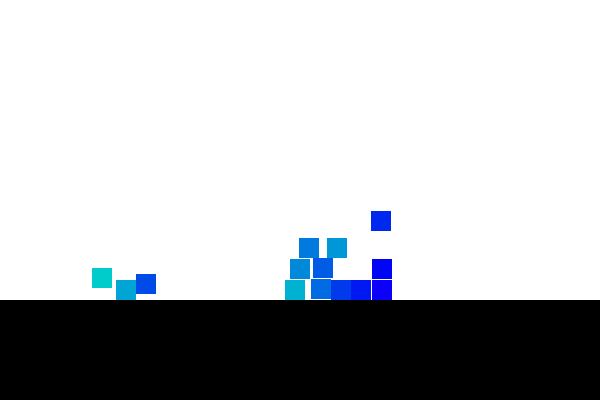}
    \caption{PairVDN, $t=1000$}
  \end{subfigure}
  \caption{Behaviour of VDN and PairVDN on Box Jump with 16 agents, no rotation, and seed zero in both cases. Notice how PairVDN's agents are grouped together more closely than VDN's at time $t=400$, and more so at time $t=1000$.}
  \label{fig:pairvdn}
\end{figure}

\Cref{tab:results_tab} gives detailed results for each model on the Box Jump task with 16 agents, with an additional column $t_\text{max}=1000$ measuring the performance on longer episodes. Note that increasing the time limit can only feasibly increase a model's score, as it has more time to build a larger tower. While \Cref{fig:marl_reward_graphs} seems to show PairVDN achieving similar performance to VDN, the $t_\text{max}=1000$ experiments reveal a significant difference: due to its non-monotonic aggregation, PairVDN leads to far more \textit{coordinated} agent behaviour, causing agents to remain clumped together for a long period of time, while they slowly disperse under VDN (\Cref{fig:pairvdn}). This is evident in the magnitude of the improvement under PairVDN when increasing $t_\text{max}$ from 400 to 1000, which is significantly larger than for any of the baselines. We consider this improved coordination to be a significant finding of our work, highlighting the weakness of monotonic value decomposition. % Under other baselines, the agents have largely scattered by $t_\text{max}=400$, leading to limited improvements beyond this point, whereas PairVDN encourages agents to remain together.

All models perform very poorly on the cooking environment. This is a challenging, complex environment, with verbose observations (dimensionality 119) and sparse rewards - further worsened by the loss of specific per-agent rewards. It is evidently too complex for our simple DQN architecture, and likely requires the use of more complex agent architectures, such as those containing recurrent units. The models additionally perform poorly on Simple Spread, achieving scores little better random, which is also likely due to the simple network architecture and small MLP size.

\newcommand{\ww}{0.32 \linewidth}
\newcommand{\imw}{0.95 \linewidth}
\begin{figure}
  \centering
  \begin{subfigure}{\ww}
    \centering
    \includegraphics[width=\imw]{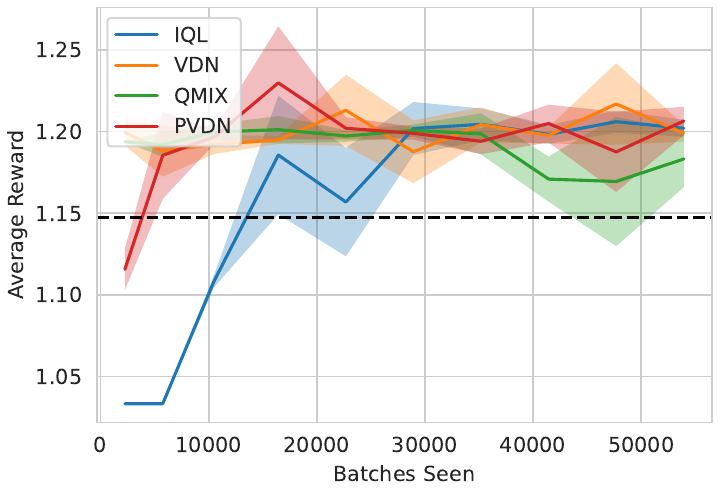}
    \caption{Box Jump, 8 agents}
  \end{subfigure}
  \begin{subfigure}{\ww}
    \centering
    \includegraphics[width=\imw]{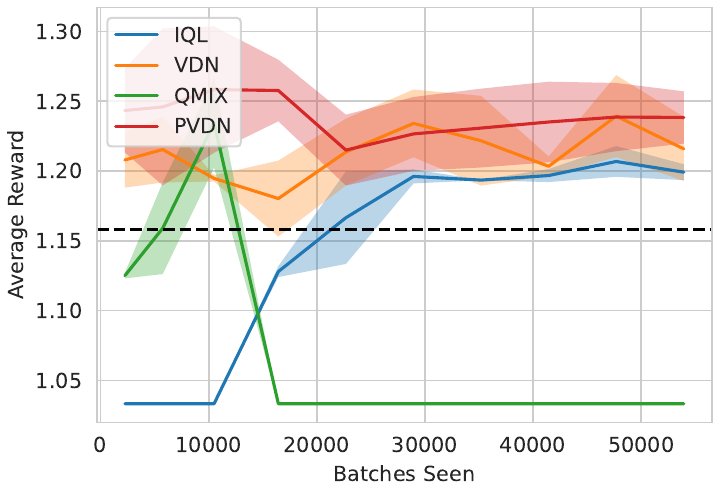}
    \caption{Box Jump, 16 agents}
  \end{subfigure}
  \begin{subfigure}{\ww}
    \centering
    \includegraphics[width=\imw]{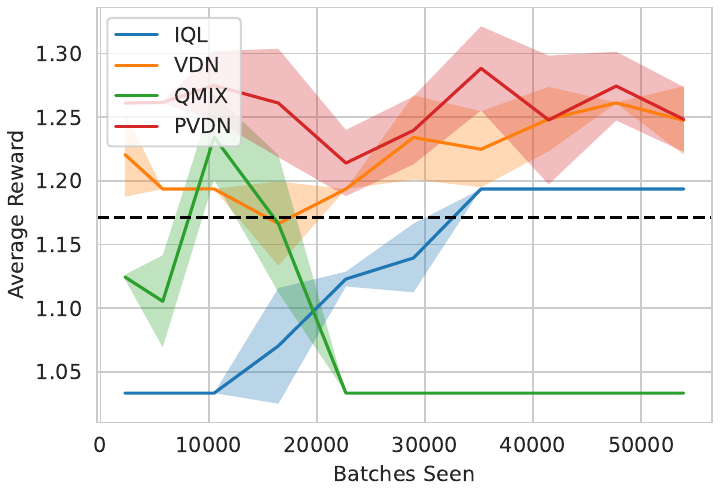}
    \caption{Box Jump, 16 agents, no rotation}
  \end{subfigure}
  \par
  \vspace{0.2cm}
  \begin{subfigure}{\ww}
    \centering
    \includegraphics[width=\imw]{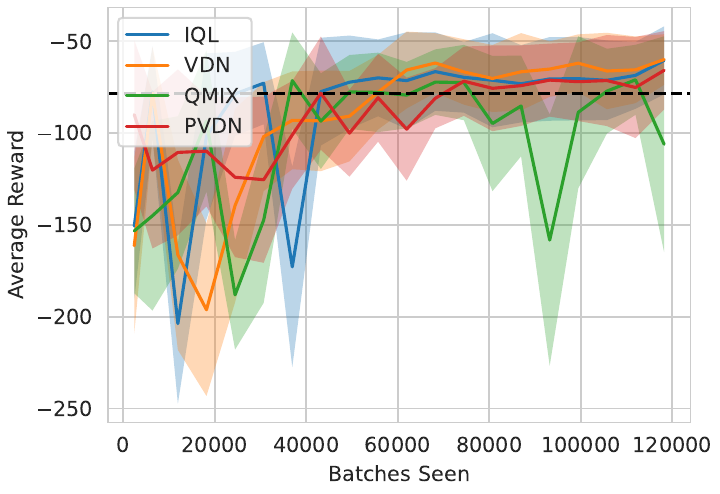}
    \caption{Simple Spread, 3 agents}
  \end{subfigure}
  \begin{subfigure}{\ww}
    \centering
    \includegraphics[width=\imw]{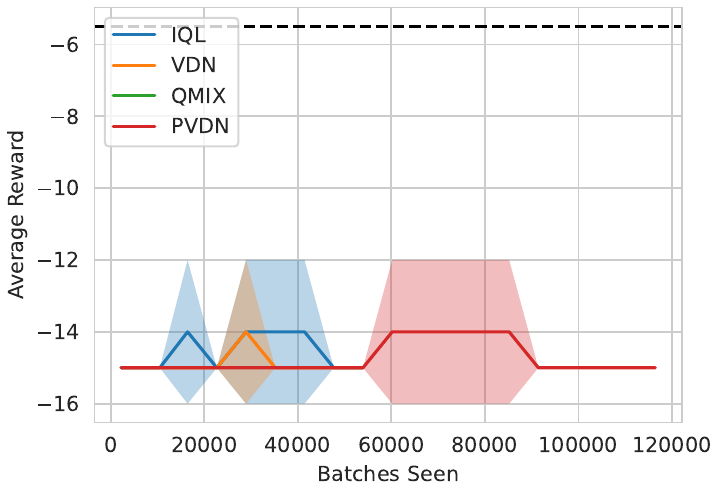}
    \caption{CookingZoo, 3 agents}
  \end{subfigure}
  \caption{Total episode rewards during training for the various trained models, with standard deviation across five episodes shown shaded. Black dashed line gives performance of a random baseline (averaged over 30 runs). PVDN denotes PairVDN.}
  \label{fig:marl_reward_graphs}
\end{figure}

% Table of final results on BOX JUMP
\begin{table}
    \centering
    \caption{Average total reward over 20 episodes on Box Jump with 16 agents after training. VDN performs similarly to PairVDN after 400 steps, but after 1000 there is a clear difference. Best mean result in bold, although the standard deviation should be taken into account.}
\begin{tabular}{c|cc|cc}
\toprule
 & \multicolumn{2}{c|}{Rotation} & \multicolumn{2}{c}{No Rotation} \\
 & \multicolumn{1}{c}{$t_\text{max}=400$} & $t_\text{max}=1000$ & \multicolumn{1}{c}{$t_\text{max}=400$} & $t_\text{max}=1000$ \\
\midrule
Random & 1.170 $\pm$ 0.025 & 1.197 $\pm$ 0.013 & 1.178 $\pm$ 0.023 & 1.225 $\pm$ 0.032 \\
IQL & 1.211 $\pm$ 0.019 & 1.212 $\pm$ 0.016 & 1.194 $\pm$ 0.000 & 1.228 $\pm$ 0.045 \\
QMIX & 1.033 $\pm$ 0.000 & 1.033 $\pm$ 0.000 & 1.033 $\pm$ 0.000 & 1.033 $\pm$ 0.000 \\
VDN & 1.224 $\pm$ 0.028 & 1.235 $\pm$ 0.028 & 1.244 $\pm$ 0.029 & 1.258 $\pm$ 0.026 \\
PairVDN & \bf 1.239 $\pm$ 0.031 & {\bf 1.271} $\pm$ 0.019 & {\bf 1.259} $\pm$ 0.026 & {\bf 1.294} $\pm$ 0.032 \\
\bottomrule
    \end{tabular}
    \label{tab:results_tab}
\end{table}

% Grouping screenshot

\section{Conclusion}
This paper proposes PairVDN, a novel decomposition of joint multi-agent value functions, which is notably able to represent \textit{non-monotonic} value functions. We further provide the non-trivial dynamic programming algorithm which is necessary to efficiently optimise PairVDN, and include a differentiable implementation in our public codebase. The terms effectively form a loop over the agents, with each agent's action accounted for by two terms, leading to improved cooperative behaviour after global maximisation. Finally, we demonstrate the efficacy of our method on a complex environment with a large number of agents, despite the simple DQN architecture, and achieve superior performance to VDN, QMIX and IQL.

% \printbibliography
% \bibliography{refs}
\bibliographystyle{plainnat}
\bibliography{main}

\appendix

\newpage
\section{PairVDN Dynamic Programming Algorithm} \label{app:dynamic_programming}
We wish to maximise $F$ where
\[
F(\acts) = \sum_{i=1}^n \Qq{i}{j}((\ob_i, \ob_j), (\act_i, \act_j)) \qquad j = \begin{cases}
    i+1 & i<n \\
    1 & i=n
\end{cases}
\]
The cyclic dependency makes this slightly more difficult, and increases the runtime of the algorithm from $\mathcal{O}(n|\A|^2)$ to $\mathcal{O}(n|\A|^3)$. Informally, we solve the non-cyclic version (without $\Qq{n}{1}$) for each possible starting action $a_1$, then use these to find the global optimum.

For $2\leq k\leq n$, define
\[
G_k(\act_1, \act_k) = \max_{\act_2,\dots,\act_{k-1}} \sum_{i=1}^{k-1} \Qq{i}{j}((\ob_i, \ob_{i+1}), (\act_i, \act_{i+1}))
\]
That is, the max sum achievable over the first $k$ terms with the given values of $a_1$ and $a_k$. Note that the upper bound on this sum is $k-1$, so $G$ never includes the `cyclic' term $\Qq{n}{1}$. It follows from this definition that
\[
G_{k+1}(\act_1, \act_{k+1}) = \max_{a_k} \bracket{G_k(\act_1, \act_k) + \Qq{k}{k+1}((\ob_k, \ob_{k+1}), (\act_k, \act_{k+1}))},
\]
and let $P_{k+1}(\act_1, \act_{k+1})$ be the corresponding argmax. When $k=2$ we have $G_2(a_1,a_2)=\Qq{1}{2}((\ob_1,\ob_2),(\act_1,\act_2))$. Using these two equations, $G$ may be implemented as an array with $n|\A|^2$ entries, and calculated completely in time $\mathcal{O}(n|\A|^3)$, the bottleneck being the iteration over $a_k$ in the equation above.

Finally, we must `close the loop' and relate $G$ back to $F$:
\[
\max_{\acts} F(\acts) = \max_{a_1, a_n}\left(G_{n}(a_1, a_n) + \Qq{n}{1}(a_n, a_1)\right)
\]
The argmax here gives the optimal values for $a_n$ and $a_1$, from which we may backtrack (using $P$) to construct the complete optimal sequence of actions.

\subsection{Future Extensions to PairVDN} \label{app:vdn_ext}
Inspired by QMIX, it may be beneficial to incorporate state-dependent weights $w_i$ for each $\Qq{i}{j}$ to improve expressivity further - in fact, these weights may be negative here, as PairVDN does not require monotonicity, unlike QMIX. The algorithm above could be easily adapted for this. 

Additionally, PairVDN is limited by its arbitrary ordering over agents, with terms covering a fixed cycle $j=i+1$. As the relevance of each agent pair is likely to change during an episode, we could instead allow the choice of $j$ for each $i$ to change at each step (for example, by choosing the agent currently closest to $i$). This changes the graph from a single directed cycle into a more complex directed structure, with $n$ nodes and $n$ edges. Such a graph is guaranteed to consist of some number of closed loops, containing $m$ nodes in total, with the remaining $n-m$ forming directed trees with their root on one of these cycles. The algorithm can be adapted for these trees - computation is faster on a tree due to the lack of cyclic dependencies, and is possible in $\mathcal{O}(n|\A|^2)$ via a similar algorithm to that above. For each tree, which is guaranteed to have a root on a cycle, the algorithm computes the best score achievable across the whole tree for each action taken at the root. Following this computation, we may run a slightly modified version of the algorithm above on each cycle in the graph individually to again retrieve the global maximum in time $\mathcal{O}(m|\A|^3+(n-m)|\A|^2)=\mathcal{O}(n|\A|^3)$.

\section{BoxJump Environment Details} \label{app:box_jump}
The action space has size $|\A|=4$, where 0 is a no-op, 1 and 2 apply forces left/right, and 3 jumps (if the box is able to jump, otherwise does nothing). Agents start each episode exactly at ground level, spaced evenly across a horizontal line, with a slight random variation in position to prevent .

The observation $\ob_i$ for each agent consists of:
\begin{itemize}
    \item Position: $(x,y)$, each ranging from 0 to 1.
    \item Velocity: $(v_x,v_y)$, given in position units per second.
    \item Angle: $\theta\in(-0.5,0.5]$, giving the number of \textit{quarter rotations}. As rotating a box by 90 degrees creates an identical state, there is no need to distinguish between these states. $\theta=0$ indicates that the box's sides are parallel to the floor, while 0.5 indicates 45 degrees of rotation.
    \item Left/right distances: the distance that a ray fired to the left/right travels before hitting another box. More precisely, this is the horizontal distance to the closest box with which this box overlaps vertically.
    \item Up/down distances: as with left/right. Down distance effectively tells the agent whether it is mid jump or stationed on the floor, while up distance tells the agent whether another agent is standing on top of it.
    \item Whether the box can currently jump.
    \item Highest $y$ value observed this episode (by any agent) up until now, $y_t^\text{best}$.
    \item The remaining time $t$. This is necessary for the environment to satisfy the Markov condition.
\end{itemize}

\begin{figure}
\centering
\begin{subfigure}{.5\textwidth}
  \centering
  \includegraphics[width=.8\linewidth]{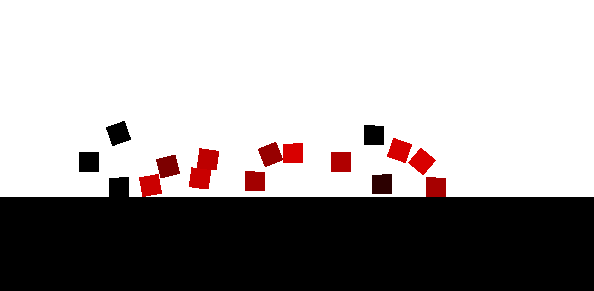}
  \caption{`Left' distance.}
\end{subfigure}%
\begin{subfigure}{.5\textwidth}
  \centering
  \includegraphics[width=.8\linewidth]{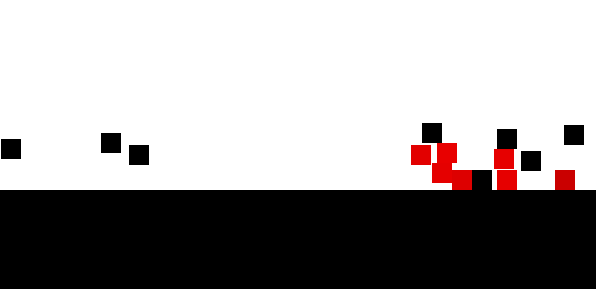}
  \caption{`Above' distance}
\end{subfigure}
\caption{Example values of the left (a) and above (b) distance observations, where red indicates a value of zero, black a value of one, with a gradient between. Notice how the `above' distance observation clearly indicates to the agent whether another agent is stacked on top of it (when the value is zero).}
\label{fig:boxenv_testing}
\end{figure}

\Cref{fig:boxenv_testing} demonstrates some example observation values visually.

We use a sparse reward signal, rewarding all agents evenly only when a new global maximum height is achieved. Formally, $r_t = y^{\text{best}}_t - y^{\text{best}}_{t-1}$, where $y^{\text{best}}_t=\max_{i,t'\leq t} y_{i,t'}$, incentivising the building of tall towers. Note that this scheme causes the total reward of an episode to be the best y value achieved over the whole episode, $\sum r_t = y^{\text{best}}_{T}$ where $T$ is the total episode length.

\subsection{Limitations}
\begin{itemize}
\item Boxes are allowed to jump iff their vertical velocity has not exceeded a small threshold in the last 15 frames. This leads to a slight delay in their jumping, and could theoretically be exploitable.
\item The observation calculation code has quadratic runtime with the number of agents due to the left/right/up/down distances - this could be improved using a more complex sweep-based algorithm.
\item Agents can walk off the sides of the map, then fall infinitely. There is a large reward penalty to handle this.
\end{itemize}

\section{Hyperparameters} \label{app:hyperparams}
\begin{table}[H]
    \centering
    \begin{tabular}{cc}
    \toprule
         Hyperparameter & Value  \\
         \midrule
         Epochs & 100\\
         Learning rate & 1e-4\\
         Batch size & 32\\
         Gamma $\gamma$ & 0.99\\
         Target network $c$ & 0.99\\
         $\epsilon$-greedy value & $1\rightarrow0.05$\\
         MLP layer sizes & $2|\St|$, 128, 128, $|\A|^2$ \\
         Exploration per epoch & 400\\
         Optimiser & SGD \\
         Experience buffer size & 20,000 \\
         \bottomrule
    \end{tabular}
    \caption{Hyperparameter values.}
    \label{tab:hyperparams}
\end{table}

\section{Additional Figures} \label{app:additional_figs}

\begin{figure}[H]
  \centering
  \begin{subfigure}{0.22\linewidth}
    \centering
    \includegraphics[width=\linewidth, trim=0cm 0cm 0cm 6cm, clip]{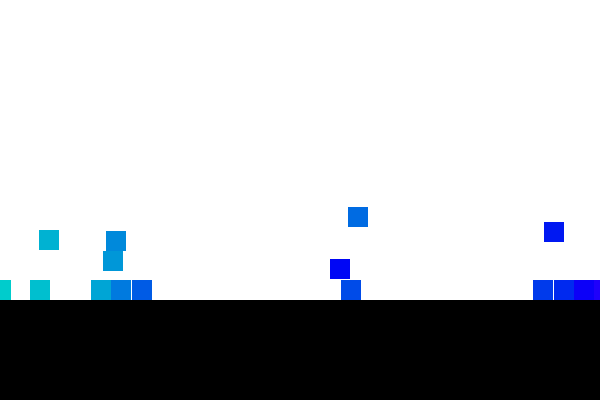}
    \caption{VDN, $t=400$}
  \end{subfigure}
  % \hfill
  \hspace{1pt}
  \begin{subfigure}{0.22\linewidth}
    \centering
    \includegraphics[width=\linewidth, trim=0cm 0cm 0cm 6cm, clip]{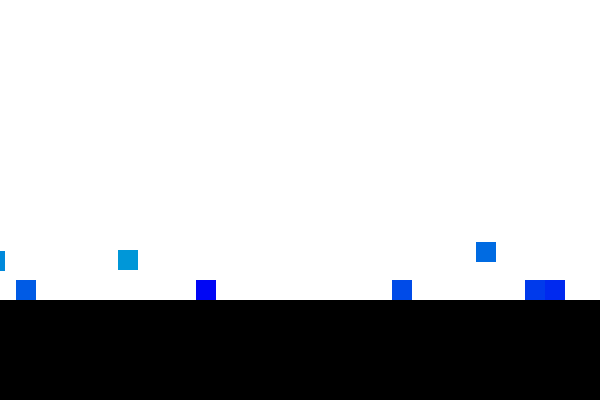}
    \caption{VDN, $t=1000$}
  \end{subfigure}
  \hfill
  \begin{subfigure}{0.22\linewidth}
    \centering
    \includegraphics[width=\linewidth, trim=0cm 0cm 0cm 6cm, clip]{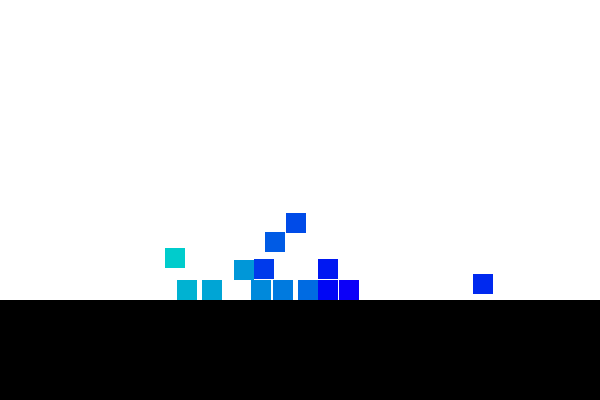}
    \caption{PairVDN, $t=400$}
  \end{subfigure}
  \hspace{1pt}
  \begin{subfigure}{0.22\linewidth}
    \centering
    \includegraphics[width=\linewidth, trim=0cm 0cm 0cm 6cm, clip]{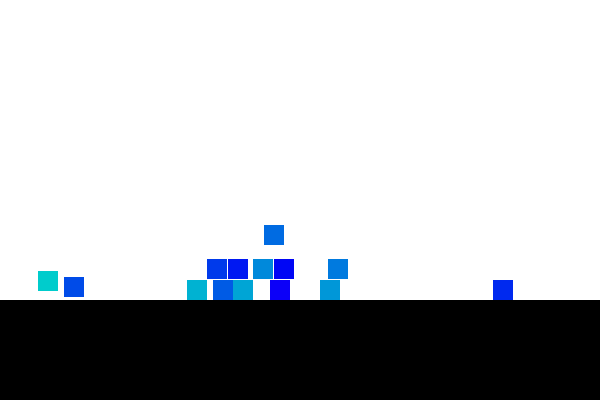}
    \caption{PairVDN, $t=1000$}
  \end{subfigure}
  \caption{\Cref{fig:pairvdn} but for seed 1 rather than seed 0, demonstrating that the figure was not cherry-picked but a frequent phenomenon.}
  \label{fig:pairvdn2}
\end{figure}

\end{document}